\pdfoutput=1
\documentclass[11pt]{article}

\usepackage{acl}

\usepackage{times}
\usepackage{latexsym}

\usepackage[T1]{fontenc}
\usepackage[utf8]{inputenc}

\usepackage{microtype}

\usepackage{inconsolata}

\usepackage{graphicx}
\usepackage{float}
\usepackage{amsfonts}
\usepackage{amsmath}
\usepackage{hyperref}
\usepackage{url}
\usepackage{graphicx}
\usepackage{multirow}
\usepackage{caption}
\usepackage{subcaption}
\usepackage{booktabs}
\usepackage{nicefrac}
\usepackage{microtype}
\usepackage{xcolor}
\usepackage{epsfig}
\usepackage{amsmath}
\usepackage{diagbox}
\usepackage{float}
\usepackage{pifont}
\usepackage{color}
\usepackage{xspace}
\usepackage{caption}
\usepackage{wrapfig}
\usepackage{varwidth}

\newcolumntype{M}{>{\begin{varwidth}{4cm}}l<{\end{varwidth}}}

\newcommand*{\rowstyle}[1]{% sets the style of the next row
  \gdef\@rowstyle{#1}%
  \@rowstyle\ignorespaces%
}
\newcolumntype{=}{% resets the row style
  >{\gdef\@rowstyle{}}%
}
\newcolumntype{+}{% adds the current row style to the next column
  >{\@rowstyle}%
}
\definecolor{revcolor}{rgb}{0,0,0}

\title{SQLformer: Deep Auto-Regressive Query Graph Generation for Text-to-SQL Translation}

\author{Adrián Bazaga, Pietro Liò, Gos Micklem \\
        University of Cambridge, Cambridge, United Kingdom \\
        \{$ar989,pl219,gm263$\}@cam.ac.uk}

\begin{document}

\newcommand{\bigparallel}{\bigm\Vert}

\maketitle
\begin{abstract}
In recent years, the task of text-to-SQL translation, which converts natural language questions into executable SQL queries, has gained significant attention for its potential to democratize data access. Despite its promise, challenges such as adapting to unseen databases and aligning natural language with SQL syntax have hindered widespread adoption. To overcome these issues, we introduce SQLformer, a novel Transformer architecture specifically crafted to perform text-to-SQL translation tasks. Our model predicts SQL queries as abstract syntax trees (ASTs) in an autoregressive way, incorporating structural inductive bias in the encoder and decoder layers. This bias, guided by database table and column selection, aids the decoder in generating SQL query ASTs represented as graphs in a Breadth-First Search canonical order. Our experiments demonstrate that SQLformer achieves state-of-the-art performance across six prominent text-to-SQL benchmarks.
\end{abstract}

\section{Introduction}

Relational databases are essential tools within various critical sectors like healthcare and industry among others. For those with technical expertise, accessing data from these databases using some form of structured query language, such as SQL, can be efficient. However, the intricate nature of SQL can make it daunting for non-technical users to learn, creating significant barriers to users.

Consequently, there has been a surge in interest in the field of text-to-SQL \cite{cai_encoder-decoder_2018, zelle_learning_1996, xu_sqlnet_2017, yu_typesql_2018, yaghmazadeh_sqlizer_2017}, which aims to convert natural language questions (NLQs) directly into SQL queries. This has the potential to dramatically reduce the obstacles faced by non-expert users when interacting with relational databases (DBs).

Early work in the field primarily focused on developing and evaluating semantic parsers for individual databases \cite{hemphill_atis_1990, dahl_expanding_1994, zelle_learning_1996, zettlemoyer_learning_2012, dong_language_2016}. However, given the widespread use of DBs, an approach based on creating a separate semantic parser for each database does not scale. 

One of the key hurdles in achieving domain generalisation \cite{wang_rat-sql_2021, cao_lgesql_2021, wang_proton_2022, cai_sadga_2022, hui_s2sql_2022} is the need for complex reasoning to generate SQL queries rich in structure. This involves the ability to accurately contextualise a user query against a specific DB by considering both explicit relations (like the table-column relations defined by the DB schema) and implicit relations (like determining if a phrase corresponds or applies to a specific column or table).

Recently, there has been a release of large-scale datasets \cite{yu_spider_2019, zhong_seq2sql_2017} comprising hundreds of DBs and their associated question-SQL pairs. This has opened up the possibility of developing semantic parsers capable of functioning effectively across different DBs \cite{guo_towards_2019, bogin_global_2019, zhang_editing-based_2019,wang_rat-sql_2021,suhr_exploring_2020,choi_ryansql_2020,Bazaga2021}. However, this demands the model to interpret queries in the context of relational DBs unseen during training, and precisely convey the query intent through SQL logic. As a result, cross-DB text-to-SQL semantic parsers cannot simply rely on memorising observed SQL patterns. Instead, they must accurately model the natural language query, the underlying DB structures, and the context of both.

Current strategies for cross-DB text-to-SQL semantic parsers generally follow a set of design principles to navigate these challenges. First, the question and schema representation are contextualised mutually by learning an embedding function conditioned on the schema \cite{hwang_comprehensive_2019,guo_towards_2019,wang_rat-sql_2021}. Second, pre-trained language models (LMs), such as BERT \cite{devlin_bert_2019} or RoBERTa \cite{liu_roberta_2019}, have been shown to greatly improve parsing accuracy by enhancing generalisation over language variations and capturing long-range dependencies. Related approaches \cite{yin_tabert_2020,yu_grappa_2021} have adopted pre-training on a BERT architecture with the inclusion of grammar-augmented synthetic examples, which when combined with robust base semantic parsers, have achieved state-of-the-art results.

In this paper, we present SQLformer, a novel Transformer variant with grammar-based decoding for text-to-SQL translation. We represent each NLQ as a graph with syntactic and part-of-speech relationships and depict the database schema as a graph of table and column metadata. Inspired by the image domain \cite{dosovitskiy_image_2021}, we incorporate learnable table and column embeddings into the encoder to select relevant tables and columns. Our model enriches the decoder input with this database information, guiding the decoder with schema-aware context. Then, the autoregressive decoder predicts the SQL query as an AST. Unlike large pre-trained language models or prompt-based techniques such as GPT-3, SQLformer offers greater efficiency and adaptability. We investigate SQLformer performance using six common text-to-SQL benchmarks of varying sizes and complexities. Our results show that SQLformer consistently achieves state-of-the-art performance across the multiple benchmarks, delivering more accurate and effective text-to-SQL capabilities on real-world scenarios.

\section{Related Work}

Earlier research often employed a sketch-based slot filling approach for SQL generation, which divides the task into several independent modules, each predicting a distinct part of the SQL query. Notable methods include SQLNet \cite{xu_sqlnet_2017}, TypeSQL \cite{yu_typesql_2018}, SQLOVA \cite{hwang_comprehensive_2019}, X-SQL \cite{he_x-sql_2019}, and RYANSQL \cite{choi_ryansql_2020}. These methods work well for simple queries but struggle with more complex scenarios typically encountered in real-world applications.

To address the challenges of complex SQL tasks, attention-based architectures have been widely adopted. For instance, IRNet \cite{guo_towards_2019} separately encodes the question and schema using a LSTM and a self-attention mechanism respectively. Schema linking is accomplished by enhancing the question-schema encoding with custom type embeddings. The SQL rule-based decoder from \cite{yin_syntactic_2017} was then used in order to decode a query into an intermediate representation, attaining a high-level abstraction for SQL.

On the other hand, graph-based approaches have also been effective in modeling complex question and database relationships. For instance, Global-GNN \cite{bogin_global_2019} models the database as a graph, while RAT-SQL \cite{wang_rat-sql_2021} introduces schema encoding and linking, attributing a relation to every pair of input items. Further developments include LGESQL \cite{cao_lgesql_2021}, which distinguishes between local and non-local relations. SADGA \cite{cai_sadga_2022} utilises contextual and dependency structure to jointly encode the question graph with the database schema graph. $\textnormal{S}^{2}\textnormal{SQL}$ \cite{hui_s2sql_2022} incorporates syntactic dependencies in a relational graph network \cite{wang_relational_2020}, and RASAT \cite{qi_rasat_2022} integrates a relation-aware self-attention module into T5 \cite{raffel_exploring_2020}. These methods have demonstrated the effectiveness of modeling questions and database schema as relational graphs.

Recent work has demonstrated the effectiveness of fine-tuning pre-trained models. For instance, \cite{shaw_compositional_2021} showed that fine-tuning a pre-trained T5-3B model could yield competitive results. Building on this, \cite{scholak_picard_2021} introduced PICARD, a technique that constrains the auto-regressive decoder by applying incremental parsing during inference time. This approach filters out grammatically incorrect sequences in real time during beam search, improving the quality of the generated SQL. RESDSQL \cite{li_resdsql_2023} proposes an schema ranking approach, retaining only the schemas most relevant to the question, before feeding it to a pre-trained RoBERTa \cite{liu_roberta_2019} in a seq2seq setting. However, these methods leverage pre-trained language models without incorporating SQL-specific constraints during decoding, which can limit their performance.

\section{Preliminaries}

\subsection{Problem Formulation}

Given a natural language question $\mathcal{Q}$ and a schema $\mathcal{S}$ = ($\mathcal{T}$, $\mathcal{C}$) for a relational database, our objective is to generate a corresponding SQL query $\mathcal{Y}$. Here, the sequence $\mathcal{Q}$ $=$ $\{$q\textsubscript{1} $\ldots$ q\textsubscript{$\vert$$\mathcal{Q}$$\vert$}$\}$ is a sequence of natural language tokens or words, where \textit{$\vert$$\mathcal{Q}$$\vert$} is the length of the question. The database schema is comprised of tables $\mathcal{T}$ $=$ $\{$t\textsubscript{1}, $\ldots$, t\textsubscript{$\vert$$\mathcal{T}$$\vert$}$\}$ and columns $\mathcal{C}$ $=$ $\{$c\textsubscript{1}, $\ldots$, c\textsubscript{$\vert$$\mathcal{C}$$\vert$}$\}$, where $\vert$$\mathcal{T}$$\vert$ and $\vert$$\mathcal{C}$$\vert$ are the number of tables and columns in the database, respectively. Each column name c\textsubscript{i} $\in$ $\mathcal{C}$, is comprised of tokens \{c\textsubscript{i,1}, $\ldots$, c\textsubscript{i,$\vert C_{i} \vert$}\} , where \textit{$\vert$C\textsubscript{i}$\vert$} is the number of tokens in the column name, and similarly table names are also comprised of tokens \{t\textsubscript{i,1}, $\ldots$, t\textsubscript{i,$\vert t_{i} \vert$}\}, where $\vert t_{i} \vert$ is the number of tokens in the table name.

\subsection{Query Construction}

We define the output SQL query $\mathcal{Y}$ as a graph, representing the AST of the query in the context-free grammar of SQL, which our model learns to generate in an autoregressive fashion. The query is an undirected graph $\mathcal{G}$ = ($\mathcal{V}$, $\mathcal{E}$), of vertices $\mathcal{V}$ and edges $\mathcal{E}$. Similar to previous works \cite{yin-neubig-2017-syntactic,wang_rat-sql_2021,qi_rasat_2022}, the nodes $\mathcal{V}$ $=$ $\mathcal{P}$ $\cup$ $\mathcal{T}$ $\cup$ $\mathcal{C}$ are the possible actions derived from SQL context-free grammar rules \cite{yin-neubig-2017-syntactic}, $\mathcal{P}$, such as \textit{SelectTable}, \textit{SelectColumn}, \textit{Root}, as well as the tables ($\mathcal{T}$) and the columns ($\mathcal{C}$) of the database schema. $\mathcal{P}$ are used to represent non-terminal nodes, depicting rules of the grammar, whereas $\mathcal{T}$ and $\mathcal{C}$ are used for terminal nodes, such as when selecting table or column names to be applied within a specific rule. The edge set $\mathcal{E}$ $=$ $\{$(v\textsubscript{i},v\textsubscript{j}) $\vert$ v\textsubscript{i}, v\textsubscript{j} $\in$ $\mathcal{V}$$\}$ defines the connectivity between the different nodes.

In particular, we choose to represent the graph using an adjacency matrix under a Breadth-First-Search (BFS) node ordering scheme, $\pi$, that maps nodes to rows of the adjacency matrix as a sequence \cite{you_graphrnn_2018}. This approach permits the modelling of graphs of varying size, such as the ones representing the ASTs of complex SQL queries. Formally, given a mapping $f_{S}$ from graph, $\mathcal{G}$, to sequences, $\mathcal{S}$, and a graph $\mathcal{G}$ with $n$ nodes under BFS node ordering $\pi$, we can formulate

\vspace{-.05in}
\begin{equation}
\mathbf{S^{\pi}} = \mathbf{f_{S}(\mathcal{G}, \pi) = (S^{\pi}_{1}, \ldots , S^{\pi}_{n})}
\end{equation}

where $S^{\pi}_{i}$ $\in$ $\{$0, 1$\}$\textsuperscript{i-1}, \textit{i} $\in$ $\{$1, $\ldots$, \textit{n}$\}$ depicts an adjacency vector between node $\pi$(v\textsubscript{i}) and the previous nodes $\pi$(v\textsubscript{j}), j $\in$ $\{$1, $\ldots$, i-1$\}$ already existing in the graph, so that:

\vspace{-.2in}
\begin{equation}
\mathbf{S^{\pi}_{i}} = \mathbf{A(^{\pi}_{1,i}, \ldots, A^{\pi}_{i-1,i})^{T}, \forall\textit{i} \in \{2, \ldots, \textit{n}\}}
\end{equation}

Then, using $S^{\pi}$, we can determine uniquely the SQL graph $\mathcal{G}$ in a sequential form and learn to predict it autoregressively.

\section{SQLformer}

\subsection{Model Overview}

\begin{figure*}[th!]
\centering
\includegraphics[width=2.0\columnwidth]{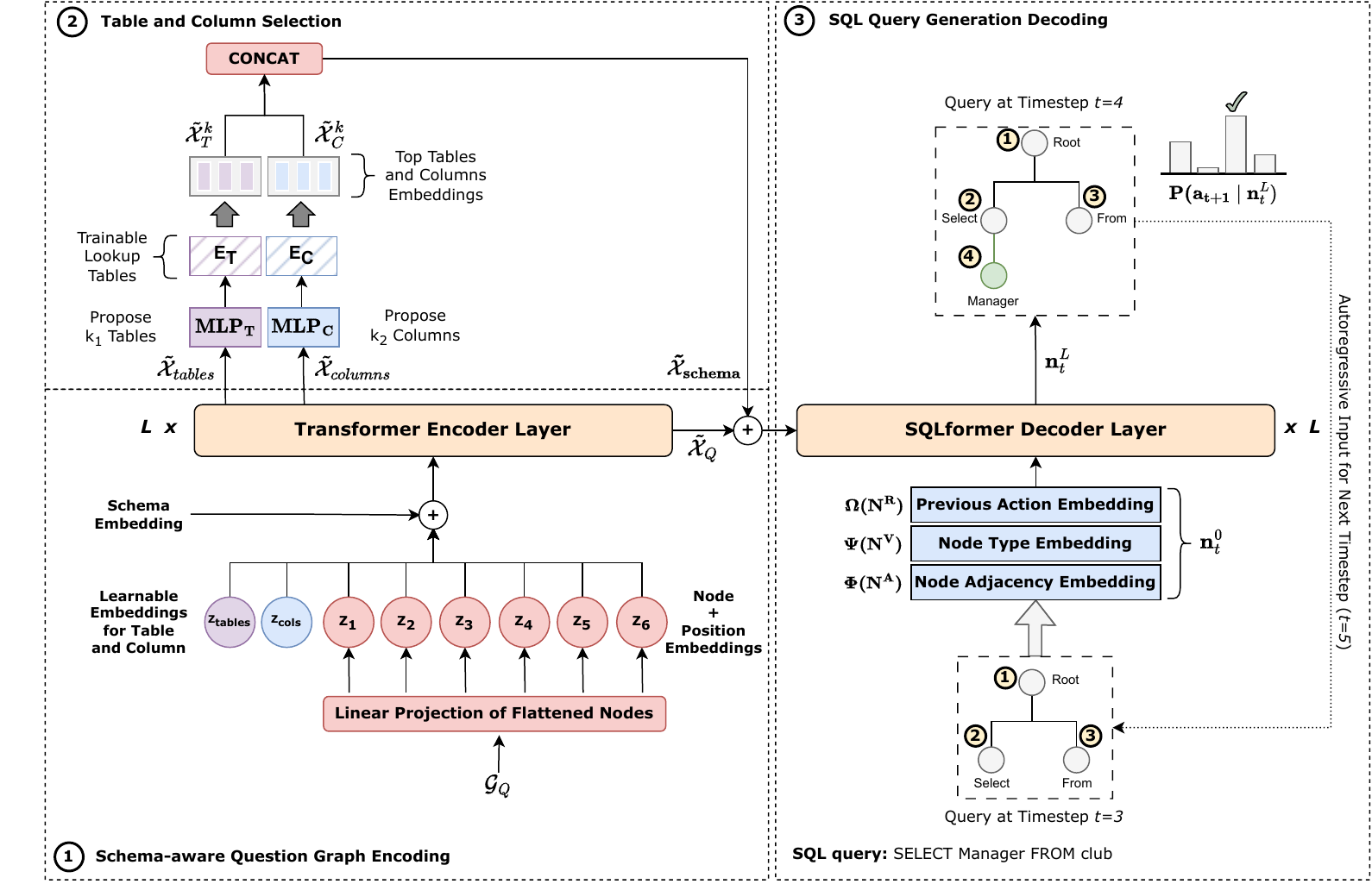}
\caption{An illustration of SQLformer: our model inherits the seq2seq nature of the Transformer architecture, consisting of $L$ layers of encoders and decoders. SQLformer encoder introduces database table and column selection as inductive biases to contextualize the embedding of a question. In this example, the question consists of six tokens (Fig. \ref{fig:figure2}). This schema-conditioned question representation serves as input to the SQLformer decoder module. Here we show the decoding timestep $t$ = 4 as an example. The architecture for the decoder module is detailed in Fig. \ref{fig:app_decoder_arch}.}
\label{fig:figure1}
\end{figure*}

In light of recent advancements \cite{shaw_compositional_2021,scholak_picard_2021,li2023graphix}, we approach the text-to-SQL problem as a translation task by using an encoder-decoder architecture. Specifically, we extend the original Transformer encoder (Subsection 4.3) by incorporating learnable table and column tokens in the encoder, used to select the most relevant tables and columns in the database schema given the NLQ. This information is injected as input to the decoder, so that it can be enriched with the representation of the schema-aware question encoding and the most relevant tables and columns in the database schema selected by the model. Moreover, the SQLformer decoder extends the original Transformer decoder (Subsection 4.4) in a way that integrates both node type, adjacency and previous generated action embeddings for generating a SQL query autoregressively as a sequence of actions derived from a SQL grammar \cite{yin-neubig-2017-syntactic}. The overall architecture of our SQLformer model is described in Fig. \ref{fig:figure1}.

\subsection{Model Inputs}

In this section, we detail how the inputs to our model are constructed, in particular, the construction of both the NLQ and schema graphs.

\begin{figure}[h]
\centering
\includegraphics[width=1.0\columnwidth]{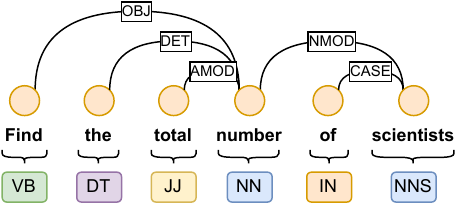}
\caption{Illustration of an example Spider question with six tokens as a graph $\mathcal{G}$ with part-of-speech and dependency relations. In this example, the token $number$ has a OBJECT dependency with $Find$, and $Find$ and $number$ are tagged as verb (VB) and noun (NN), respectively.}
\vspace{-.1in}
\label{fig:figure2}
\end{figure}

\paragraph{Question Graph Construction.} The natural language question can be formulated as a graph $\mathcal{G}_{Q}$ = ($\mathcal{Q}$, $\mathcal{R}$), where the node set $\mathcal{Q}$ consists of the natural language tokens, and $\mathcal{R}$ = $\{$r\textsubscript{1}, $\ldots$, r\textsubscript{$\vert$$\mathcal{R}$$\vert$}$\}$ represents one-hop relations between words. We employ two types of relations for the question graph: syntactic dependencies and part-of-speech tagging, incorporating grammatical meaning. These relations form a joint question graph, which is then linearized as a Levi graph. Fig. \ref{fig:figure2} illustrates an example question graph with some relationships. Tables \ref{tab:pos_tags} and \ref{tab:dep_tags} describe all relations used. To encode each token in the question graph, we use a Graph Attention Network (GAT) \cite{velickovic_graph_2018}.

\begin{figure}[ht]
\centering
\includegraphics[width=1.0\columnwidth]{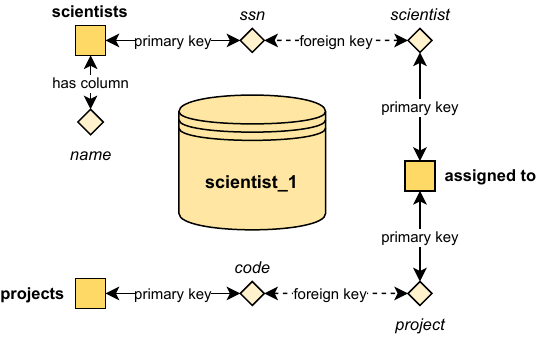}
\caption{An illustration of an example Spider schema for database $scientist\_1$. In this example, there are a total of 3 tables ($scientists$, $projects$, $assigned\_to$), with multiple columns for each table and relationships between the tables.}
\vspace{-.1in}
\label{fig:figure3}
\end{figure}

\paragraph{Database Schema Graph Construction.} Similarly, a database schema graph is represented by $\mathcal{G}_{S}$ = ($\mathcal{S}$, $\mathcal{R}$) where the node set $\mathcal{S}$ $=$ ($\mathcal{T}$, $\mathcal{C}$) represents the tables, $\mathcal{T}$, and the columns, $\mathcal{C}$, in the schema. The edge set $\mathcal{R}$ $=$ $\{$r\textsubscript{1}, $\ldots$, r\textsubscript{$\vert$$\mathcal{R}$$\vert$}$\}$ depicts the structural relationships among tables and columns in the schema. Similarly to previous works, we use the common relational database-specific relations, such as primary/foreign key for column pairs, column types, and whether a column belongs to a specific table. Fig. \ref{fig:figure3} shows an example database schema graph and Table \ref{tab:schema_relations} provides a description of the types of relationships used for database schema graph construction. We encode the schema using a GAT and use average pooling to obtain a single embedding to represent each database schema.

\subsection{Table and Column Selection Encoder}

To describe our proposed modification to the Transformer encoder, we first introduce the original Transformer architecture. The Transformer encoder \cite{vaswani_attention_2017} consists of alternating layers of multi-head self-attention (MHA) and Fully-connected Forward Network (FFN) blocks. Before every block, Layer Normalisation (LN) is applied, and after every block, a residual connection is added. More formally, in the $\ell^{th}$ encoder layer, the hidden states are represented as $X^{\ell}_{S} = \{x^{\ell}_{1}, \ldots, x^{\ell}_{N}\}$, where \textit{N} is the maximum length of the inputs. 

First, a MHA block maps \textit{X} into a query matrix \textit{Q} $\in$ $\mathbb{R}^{n\times d_{k}}$, key matrix \textit{K} $\in$ $\mathbb{R}^{n\times d_{k}}$ and value matrix \textit{V} $\in$ $\mathbb{R}^{n\times d_{v}}$, where \textit{m} is the number of query vectors, and \textit{n} the number of key or value vectors. Then, an attention vector is calculated as follows:

\vspace{-.2in}
\begin{equation}
\begin{aligned}
\mathbf{Attention(Q, K, V)} {=} & \mathbf{softmax(A)} \mathbf{V},\\
\mathbf{A} {=} & \frac{\mathbf{Q} \mathbf{K^T}}{\mathbf{\sqrt{d_{k}}}}
\end{aligned}
\label{eqn:attn_a_matrix}
\end{equation}

In practice, the MHA block calculates the self-attention over \textit{h} heads, where each head \textit{i} is independently parametrized by $\mathbf{W^{Q}_{i}}$ $\in$ $\mathbb{R}^{d_{m}\times d_{k}}$, $\mathbf{W^{K}_{i}}$ $\in$ $\mathbb{R}^{d_{m}\times d_{k}}$ and $\mathbf{W^{V}_{i}}$ $\in$ $\mathbb{R}^{d_{m}\times d_{v}}$, mapping the input embeddings $\mathcal{X}$ into queries and key-value pairs. Then, the attention for each head is calculated and concatenated, as follows:

\vspace{-.2in}
\begin{equation}
\begin{aligned}
\mathbf{Head_{i}} {=} & \mathbf{Attention(Q W^{Q}_{i}, K W^{K}_{i}, V W^{V}_{i})}\\
\mathbf{MHA(\mathcal{X}^{\ell}_\mathcal{S})} {=} & \mathbf{Concat(Head_{1}, \ldots, \textnormal{Head}_{h}) W^{U}}\\
\mathbf{\bar{\mathcal{X}}^{\ell}_\mathcal{S}} {=} & \mathbf{MHA(\mathcal{X}^{\ell}_\mathcal{S})}
\end{aligned}
\end{equation}

where $\mathbf{W^{U}}$ $\mathbf{\in}$ $\mathbb{R}^{d^{h}_{m}\times d_{m}}$ is a trainable parameter matrix. Next, to acquire the hidden states of the input, a FFN block is applied, as follows:

\vspace{-.2in}
\begin{equation}
\mathbf{FFN(\bar{\mathcal{X}}^{\ell}_\mathcal{S})} = \mathbf{max(0, \bar{\mathcal{X}}^{\ell}_\mathcal{S} W_{1} + b_{1}) W_{2} + b_{2}}
\end{equation}

where $\mathbf{W_{1}}$ $\in$ $\mathbb{R}^{d_{m}\times d_{ff}}$ and $\mathbf{W_{2}}$ $\in$ $\mathbb{R}^{d_{ff}\times d_{m}}$ are linear weight matrices. Finally, layer normalisation and residual connection are applied as follows:

\vspace{-.2in}
\begin{equation}
\mathbf{\tilde{\mathcal{X}}^{\ell}_\mathcal{S} = \mathbf{LayerNorm}(\bar{\mathbf{\mathcal{X}}}^{\ell}_\mathcal{S} + \mathbf{FFN}(\bar{\mathbf{\mathcal{X}}}^{\ell}_\mathcal{S}))}
\end{equation}

In the SQLformer encoder, we input the 1D sequence of natural language token embeddings, $Z$, and prepend two learnable tokens: $Z_{tables}$ and $Z_{cols}$. The states of these tokens at the encoder output, $\tilde{\mathcal{X}}_{tables}$ and $\tilde{\mathcal{X}}_{columns}$, serve as input to two MLP blocks responsible for selecting $k_{1}$ tables and $k_{2}$ columns based on the NLQ. Sinusoidal vectors retain the original positional information.

After \textit{L} encoder layers, we obtain the input question embedding $\tilde{\mathcal{X}}^{\ell}_\mathcal{S}$, with the first two tokens as $\tilde{\mathcal{X}}_{tables}$ and $\tilde{\mathcal{X}}_{columns}$, and the rest as natural language question tokens $\tilde{\mathcal{X}}_{Q}$ $\in$ $\mathbb{R}^{d\times Q}$. $\tilde{\mathcal{X}}_{T}$ and $\tilde{\mathcal{X}}_{C}$ input to MLP blocks $\mathbf{MLP_{T}}$ $\in$ $\mathbf{\mathbb{R}^{d\times \vert \mathcal{T} \vert}}$ and $\mathbf{MLP_{C}}$ $\in$ $\mathbf{\mathbb{R}^{d\times \vert \mathcal{C} \vert}}$, where $d$ is the hidden size of the token embeddings, and $\vert \mathcal{T} \vert$ and $\vert \mathcal{C} \vert$ are the sizes of the tables and columns vocabularies, respectively. The embeddings are projected into probability vectors:

\vspace{-.15in}
\begin{equation}
\begin{aligned}
\mathbf{P_{tables}} {=} & \mathbf{softmax(MLP_{T}(\tilde{\mathcal{X}}_{T}))}\\
\mathbf{P_{columns}} {=} & \mathbf{softmax(MLP_{C}(\tilde{\mathcal{X}}_{C}))}
\end{aligned}
\end{equation}

Then, the top $k_{1}$ and $k_{2}$ tables and columns, respectively, are selected according to $\mathbf{P_{tables}}$ and $\mathbf{P_{columns}}$. A masking vector is applied to $\mathbf{P_{columns}}$ to ensure that only columns from the selected tables are considered, avoiding the selection of columns not present in the selected tables. Next, two embedding lookup tables, $\mathbf{E_{T}}$ $\in$ $\mathbf{\mathbb{R}^{\vert \mathcal{T} \vert \times d_{t}}}$ and $\mathbf{E_{C}}$ $\in$ $\mathbf{\mathbb{R}^{\vert \mathcal{C} \vert \times d_{c}}}$, are used for mapping the \textit{k} top tables and columns, respectively, into embeddings, as $\tilde{\mathcal{X}}^{k}_{T}$ $\in$ $\mathbf{\mathbb{R}^{k_{1}\times d}}$ and $\tilde{\mathcal{X}}^{k}_{C}$ $\in$ $\mathbf{\mathbb{R}^{k_{2}\times d}}$, where $d$ is the size of the learnable embeddings. These are aggregated and concatenated, giving the final representation for the schema, depicted as $\mathbf{\tilde{\mathcal{X}}_{schema}}$

Finally, $\tilde{\mathcal{X}}_{Q}$ and $\mathbf{\tilde{\mathcal{X}}_{schema}}$ are aggregated to effectively contextualize the natural language question embedding by the embedding of the most likely tables and columns in the schema being mentioned. The result of this aggregation is given as input to the decoder module as part of the cross-attention.

\subsection{Autoregressive Query Graph Generation Decoder}

During the decoding phase, previous works (e.g. \citet{wang_rat-sql_2021, cao_lgesql_2021, hui_s2sql_2022, cai_sadga_2022}) widely adopt the LSTM-based tree decoder from \citet{yin_syntactic_2017} to generate SQL grammar rules. In contrast, the SQLformer decoder (Fig. \ref{fig:app_decoder_arch}) extends the original Transformer decoder to predict the SQL grammar rules autoregressively. This approach has multiple advantages. First, it maintains the context of previously generated parts of the query for longer sequences than LSTM-based decoders. This is especially important for long queries, such as these containing sub-queries. Also, the Transformer encourages permutation invariance desirable for processing the node embeddings of the SQL graph, as the graph is invariant under any permutation of the nodes. Additionally, the highly parallelizable nature of the inherited Transformer architecture results in higher efficiency for both training and inference speed compared to previous LSTM-based approaches (see Table \ref{tab:table_tr_in_time} for an analysis on training and inference efficiency).

In the SQLformer decoder, each query node is described by three attributes: node type, node adjacency, and the previous action. Nodes are assigned a type, represented as $N^{V}$ $=$ $\{$V\textsubscript{0}, V\textsubscript{1}, $\ldots$, V\textsubscript{N}$\}$, where $V_{i}$ is a one-hot representation of the node type. Nodes are grouped as non-terminal or terminal, with terminal types including $table_id$ and $column_id$. Node type embeddings are calculated using a learnable transformation $\mathbf{\Psi}(N^{V})$ $\in$ $\mathbf{\mathbb{R}^{\vert V \vert \times d_{V}}}$, where $d_{V}$ is the embedding dimensionality and $\vert$V$\vert$ is the number of possible node types. Node adjacency is represented as $N^{A}$ $=$ $\{$A\textsubscript{0}, A\textsubscript{1}, $\ldots$, A\textsubscript{N}$\}$, with $A_{i}$ $\in$ $\{$0, 1$\}$\textsuperscript{M}, and embeddings obtained from $\mathbf{\Phi}(N^{A})$ $\in$ $\mathbf{\mathbb{R}^{1 \times d_{A}}}$, with $d_{A}$ as the embedding dimensionality. The previous action embedding, $a_{t-1}$, is given by the transformation $\mathbf{\Omega}(N^{R})$ $\in$ $\mathbf{\mathbb{R}^{1 \times d_{T}}}$, where $N^{R}$ is the SQL grammar rule chosen in the previous timestep and $d_{T}$ is the embedding dimensionality.

We extend the Transformer decoder architecture to incorporate the node type, adjacency and previous action embeddings to represent a node at each timestep. In particular, inspired by \cite{ying_transformers_2021}, we include the node type and adjacency embeddings in the multi head self-attention aggregation process as a bias term (see Fig. \ref{fig:app_decoder_arch} for an illustration). Formally, we modify Eq. \ref{eqn:attn_a_matrix} so that $\mathbf{\Psi}(N^{V})$ and $\mathbf{\Phi}(N^{A})$ act as a bias term in the attention calculation, as follows

\vspace{-.1in}
\begin{equation}
\begin{aligned}
\mathbf{A} {=} & \mathbf{\frac{Q K^{T}}{\sqrt{d_{k}}} + U}\\
\mathbf{U} {=} & \mathbf{\Psi(N^{V})} + \mathbf{\Phi(N^{A})}
\end{aligned}
\end{equation}

Then, the updated residuals for the node embedding, $\mathbf{n}^{\ell}_{t}$, at layer $\ell$, can be formalised as

\vspace{-.2in}
\begin{equation}
\begin{aligned}
  \mathbf{n}^{\ell}_{t} {=} & \mathbf{n^{\ell-1}_{t} + \mathbf{O}^\ell \bigparallel_{k=1}^{K} \sum_{j=1}^{N} \left(G^{k,\ell} \textnormal{ } \mathbf{V}^{k,\ell}\right)}\\
  \mathbf{G^{k,\ell}} {=} & \mathbf{softmax(A^{k,\ell})}
\end{aligned}
\end{equation}

where $\parallel$ means concatenation, and \textit{K} is the number of attention heads. As a result, the decoder state at the current timestep after $L$ decoder layers, $\mathbf{n}^{L}_{t}$, is fed to an action output MLP head which computes the distribution $P(a_{t+1})$ of next timestep actions based on the node type, adjacency, and previous action at timestep $t$. Formally, $P(a_{t+1})$ is calculated as follows

\vspace{-.2in}

\begin{equation}
\mathbf{P(a_{t+1}} \mid \mathbf{n}^{L}_{t}) = \mathbf{softmax(W_{a} n^{L}_{t})}
\end{equation}

Finally, the prediction of the SQL query AST can be decoupled into a sequence of actions $a$ = ($a_1$, $\ldots$ , $a_{|a|}$), yielding the training objective for the task as

\vspace{-.2in}

\begin{equation}
\mathbf{\mathcal{L}} = \mathbf{-\sum_{p=1}^{|a|} log \textnormal{ } P(a_{p} \mid a_{<p}, \mathcal{S}, \mathcal{Q})}
\end{equation}

\section{Experiments}

In this section, we show our model performance on six common text-to-SQL datasets. Also, we present ablation studies to analyse the importance of the different components of the SQLformer architecture.

\begin{table*}[th!]
\centering
\small
\setlength{\tabcolsep}{5mm}{
\small
\resizebox{\textwidth}{!}{
\begin{tabular}{=l|cc|cc}
\hline
\multirow{2}{*}{\textbf{Method}}     & \multicolumn{2}{c|}{\textbf{EM}} & \multicolumn{2}{c}{\textbf{EX}} \\ 
\cline{2-5}
& \textbf{Dev} & \textbf{Test} & \textbf{Dev} & \textbf{Test}  \\
\hline
SADGA + GAP \cite{cai_sadga_2022} & 73.1 & 70.1 & - & - \\ 
RAT-SQL + GraPPa \cite{yu_grappa_2021} & 73.4 & 69.6 & - & - \\ 
RAT-SQL + GAP + NatSQL \cite{Shi_2021} & 73.7 & 68.7 & 75.0 & 73.3 \\
SMBOP + GraPPa \cite{rubin-berant-2021-smbop} & 74.7 & 69.5 & 75.0 & 71.1 \\
DT-Fixup SQL-SP + RoBERTa \cite{xu-etal-2021-optimizing} & 75.0 & 70.9 & - & -\\
LGESQL + ELECTRA \cite{cao_lgesql_2021} & 75.1 & 72.0 & - & -\\ 
RASAT \cite{qi_rasat_2022} & 75.3 & 70.9 & 80.5 & 75.5\\ 
T5-3B \cite{scholak_picard_2021} & 75.5 & 71.9 & 79.3 & 75.1\\ 
S\textsuperscript{2}SQL + ELECTRA \cite{hui_s2sql_2022} & 76.4 & 72.1 & - & -\\ 
RESDSQL \cite{li_resdsql_2023} & 80.5 & 72.0 & 84.1 & \underline{79.9} \\
GRAPHIX-T5-3B \cite{li2023graphix} & 77.1 & \underline{74.0} & 81.0 & 77.6 \\
\hline
\textbf{SQLformer} (our approach) & \textbf{78.2} & \textbf{75.6} & \textbf{82.5} & \textbf{81.9}\\ 
\hline
\end{tabular}}}
\caption{EM and EX results on Spider's dev and test dataset splits. We compare our approach with recent state-of-the-art methods. Underline depicts the previous best performing method for each metric.}
\vspace{-.1in}
\label{tab:table1}
\end{table*}

\subsection{Experimental Setup}

\paragraph{Dataset.} We consider six benchmark datasets, with complete details included in Appendix \ref{appendix:datasets}. In particular, our experiments use the Spider \cite{yu_spider_2019} dataset, a large-scale cross-domain text-to-SQL benchmark, as well as context-dependent benchmarks such as the SparC \cite{yu2019sparc} and CoSQL \cite{yu2019cosql} datasets. Additionally, we evaluate our method for zero-shot domain generalization performance on the Spider-DK \cite{gan2021exploring}, Spider-SYN \cite{gan2021robustness} and Spider-Realistic datasets. 

\paragraph{Evaluation Metrics.} We report results using the same metrics as previous works \cite{yu_spider_2019, li_resdsql_2023, li2023graphix}. For Spider-family datasets (i.e. Spider, Spider-DK, Spider-SYN and Spider-Realistic), we consider two prevalent evaluation metrics: Exact Match (EM) and Execution (EX) accuracies. The EX metric evaluates whether the predicted and ground-truth SQL queries yields the same execution results on the database. However, there can be instances where EX gives false positives. To counteract this, EM evaluates how much a predicted SQL query is comparable to the ground truth query. For SParC and CoSQL, we measure EM at the question (QEM) and interaction (IEM) levels, as well as EX at both question (QEX) and interaction levels (IEX).

\paragraph{Implementation Details.} We implemented SQLformer in PyTorch \cite{paszke_pytorch_2019}. For the graph neural network components, we use PyTorch Geometric \cite{fey_fast_2019}. The questions, column and table names are tokenized and lemmatized using \textit{stanza} \cite{qi_stanza_2020}. For dependency parsing and part-of-speech tagging, \textit{stanza} \cite{qi_stanza_2020} is used. We find the best set of hyperparameters on a randomly sampled subset of 10\% queries from the dev dataset. For training, we set the maximum input length as 1024, maximum number of generated AST nodes to 200, batch size as 32 and maximum training steps to 20,000. A detailed list of hyperparameters can be found in Appendix \ref{appendix:hyperparameter_list}. Tokens embeddings are initialized with ELECTRA \cite{clark2020electra} using the HuggingFace library \cite{wolf_transformers_2020}. We use teacher forcing in the decoder for higher training stability. Results are on the test set unless stated otherwise.

\subsection{Overall Performance}

\paragraph{Results on Spider.} The EM and EX accuracy results on the Spider benchmark are presented in Table \ref{tab:table1}. Our proposed model SQLformer achieves competitive performance in both EM and EX accuracy. On the test set, compared with RAT-SQL \cite{wang_rat-sql_2021}, our model’s EM increases from 69.6\% to 75.6\%, achieving a 6.0\% absolute improvement. When compared to approaches that fine-tune a Language Model (LM) with a much larger amount of parameters, such as T5-3B (71.9\%), we achieve a 3.7\% absolute improvement. This effectively shows the benefit of our proposed architecture for solving text-to-SQL tasks with fewer parameters. Furthermore, SQLformer sets a new state-of-the-art in EX accuracy with 81.9\%. Compared to RESDSQL \cite{li_resdsql_2023}, which achieves 72.0\% EM and 79.9\% EX, SQLformer surpasses it by 3.6\% and 2.0\% respectively. Similarly, SQLformer outperforms GRAPHIX-T5 \cite{li2023graphix}, which has 74.0\% EM and 77.6\% EX, by 1.6\% and 4.3\%. Against other methods like RASAT, SQLformer shows significant improvements of 4.7\% in EM and 6.4\% in EX. These comparisons highlight the effectiveness of SQLformer in generating highly accurate SQL queries, significantly improving upon existing state-of-the-art methods.

\paragraph{Results on Difficult Queries.} We provide a breakdown of accuracy by query difficulty level (easy, medium, hard, extra hard) as defined by \citet{yu_spider_2019}. Table \ref{tab:table2} compares our approach to state-of-the-art baselines on the EM accuracy metric. Performance drops with increasing query difficulty, from 92.7\% on $easy$ to 51.2\% on $extra$ hard queries. Compared to RAT-SQL, SQLformer shows improvements of 9.7\% on $hard$ and 8.3\% on $extra$ hard queries, demonstrating its effectiveness in handling complex queries. Therefore, SQLformer surpasses the baseline methods across all four subsets by a significant margin, giving supporting evidence for the effectiveness of our approach.

\begin{table}[h]
\resizebox{1\columnwidth}{!}{
\begin{tabular}{l|ccccc}
\hline
\textbf{Method}     & \textbf{Easy} & \textbf{Medium} & \textbf{Hard} & \textbf{Extra} & \textbf{All} \\ 
\hline
RAT-SQL + BERT & 86.4 & 73.6 & 62.1 & 42.9 & 69.7 \\ 
SADGA & 90.3 & 72.4 & 63.8 & 49.4 & 71.6 \\ 
LGESQL & 91.5 & 76.7 & 66.7 & 48.8 & 74.1 \\
GRAPHIX-T5-3B & 91.9 & 81.6 & 61.5 & 50 & 75.6 \\ 
\hline
\textbf{SQLformer} (our approach) & \textbf{92.7} & \textbf{82.9} & \textbf{71.8} & \textbf{51.2} & \textbf{76.8} \\ 
\hline
\end{tabular}}
\caption{EM accuracy on the Spider queries across different levels of difficulty as defined by \citet{yu_spider_2019}.}
\vspace{-.1in}
\label{tab:table2}
\end{table}

\paragraph{Zero-Shot Results on Domain Generalization and Robustness.} In Table \ref{tab:table_syn_dk_r}, we analyze SQLformer's capabilities in zero-shot domain generalization and robustness on the Spider-DK, Spider-SYN, and Spider-Realistic benchmarks. SQLformer excels with EM accuracies of 55.1\% on Spider-DK, 71.2\% on Spider-SYN, and 78.7\% on Spider-Realistic. These results surpass models like LGESQL with ELECTRA and sophisticated systems like GRAPHIX-T5-3B by 3.9\%, 4.3\%, and 6.3\% on DK, SYN, and Realistic, respectively, and RESDSQL by 1.8\%, 2.1\%, and 1.3\%. SQLformer's EX accuracies of 68.2\%, 78.4\%, and 82.6\% also outperform RESDSQL, demonstrating SQLformer's ability to adapt to unseen domains without direct prior training potential for real-world applications where database schemas and linguistic variations are highly variable.

\begin{table}[h]
\resizebox{1\columnwidth}{!}{
\begin{tabular}{=l|+c+c|+c+c|+c+c}

\hline
\multirow{2}{*}{\textbf{Method}}     & \multicolumn{2}{c|}{\textbf{Spider-DK}} & \multicolumn{2}{c}{\textbf{Spider-SYN}} & \multicolumn{2}{c}{\textbf{Spider-R}} \\ 
\cline{2-7}
& \textbf{EM} & \textbf{EX} & \textbf{EM} & \textbf{EX} & \textbf{EM} & \textbf{EX}  \\
\hline
RAT-SQL + GraPPa \cite{yu_grappa_2021} & 38.5 & - & 49.1 & - & 59.3 & - \\ 
LGESQL + ELECTRA \cite{cao_lgesql_2021} & 48.4 & - & 64.6 & - & 69.2 & - \\
T5-3B \cite{scholak_picard_2021} & - & - & - & - & 68.7 & 71.4 \\
GRAPHIX-T5-3B \cite{li2023graphix} & 51.2 & - & 66.9 & - & 72.4 & - \\
RESDSQL \cite{li_resdsql_2023} & 53.3 & 66.0 & 69.1 & 76.9 & 77.4 & 81.9 \\
\hline
\textbf{SQLformer} (our approach) & \textbf{55.1} & \textbf{68.2} & \textbf{71.2} & \textbf{78.4} & \textbf{78.7} & \textbf{82.6} \\ 
\hline
\end{tabular}}
\caption{EM and EX on Spider-SYN, Spider-DK and Spider-Realistic benchmarks.}
\vspace{-.1in}
\label{tab:table_syn_dk_r}
\end{table}

\paragraph{Results on Context-Dependent Settings.} We present the experimental results for SQLformer in comparison with several leading methods on the SParC (Table \ref{tab:results_sparc}) and CoSQL (Table \ref{tab:results_cosql}) datasets. For the SParC dataset, SQLformer achieves 68.6\% QEM and 51.3\% IEM, outperforming RASAT by 1.9\% and 4.1\% respectively. Additionally, SQLformer shows significant improvements in QEX and IEX metrics, with 74.5\% and 55.8\%, further confirming its superior capacity for maintaining contextual understanding in multi-turn SQL dialogue tasks. For the CoSQL dataset, SQLformer attains 60.2\% QEM and 31.4\% IEM, surpassing RASAT by 1.4\% and 5.1\%. Moreover, SQLformer’s QEX and IEX scores are 68.4\% and 39.2\%, respectively, highlighting its potential in enhancing interactive SQL query generation. These results underscore the effectiveness of SQLformer in delivering more accurate and contextually aware SQL interpretations compared to previous leading methods.

\begin{table}[h]
\renewcommand\arraystretch{1.}
\centering
\setlength{\tabcolsep}{3mm}{
\small
\resizebox{1\columnwidth}{!}{
\begin{tabular}{=l|cccc}
\hline
\textbf{Method} & \textbf{QEM} & \textbf{IEM} & \textbf{QEX} & \textbf{IEX}    \\
\hline
EditSQL + BERT \cite{zhang_editing-based_2019} & 47.2 & 29.5 & - & - \\ 
IGSQL + BERT \cite{cai-wan-2020-igsql} & 50.7 & 32.5 & - & - \\ 
RAT-SQL + SCoRe \cite{yu2021score} & 62.2 & 42.5 & - & - \\
RASAT \cite{qi_rasat_2022} & 66.7 & 47.2 & 72.5 & 53.1 \\
\hline
\textbf{SQLformer} (our approach) & \textbf{68.6} & \textbf{51.3} & \textbf{74.5} & \textbf{55.8} \\ 
\hline
\end{tabular}}}
\caption{Evaluation results on the SParC dataset.}
\vspace{-.1in}
\label{tab:results_sparc}
\end{table}

\begin{table}[h]
\renewcommand\arraystretch{1.}
\centering
\setlength{\tabcolsep}{3mm}{
\small
\resizebox{1\columnwidth}{!}{
\begin{tabular}{=l|cccc}
\hline
\textbf{Method} & \textbf{QEM} & \textbf{IEM} & \textbf{QEX} & \textbf{IEX}    \\
\hline
EditSQL + BERT \cite{zhang_editing-based_2019} & 39.9 & 12.3 & - & - \\ 
IGSQL + BERT \cite{cai-wan-2020-igsql} & 44.1 & 15.8 & - & - \\ 
RAT-SQL + SCoRe \cite{yu2021score} & 52.1 & 22.0 & - & - \\
T5-3B \cite{scholak_picard_2021} & 56.9 & 24.2 & - & - \\
HIE-SQL + GraPPa \cite{zheng2022hiesql} & 56.4 & 28.7 & - & - \\ 
RASAT \cite{qi_rasat_2022} & 58.8 & 26.3 & 66.7 & 37.5 \\
\hline
\textbf{SQLformer} (our approach) & \textbf{60.2} & \textbf{31.4} & \textbf{68.4} & \textbf{39.2} \\ 
\hline
\end{tabular}}}
\caption{Evaluation results on the CoSQL dataset.}
\vspace{-.1in}
\label{tab:results_cosql}
\end{table}

\subsection{Ablation Study}

To validate the importance of each component in our architecture, we performed ablation studies on the SQLformer model. Table \ref{tab:table5} compares the impact of four critical design choices: removing table and column selection, part-of-speech question encoding, and dependency graph question encoding. Additionally, we analyze the impact of varying the number of selected tables ($k_1$) and columns ($k_2$) on the performance of SQLformer (see Table \ref{tab:table_impact_k1_k2}).

\begin{table}[H]
\resizebox{1\columnwidth}{!}{
\begin{tabular}{l|c}
\hline
\textbf{Method}     & \textbf{EM accuracy ($\%$)} \\ 
\hline
SQLformer w/o table + column selection & 72.3 $\pm$ 0.38 \\ 
SQLformer encoder + LSTM-based decoder & 74.2 $\pm$ 0.38 \\ 
SQLformer w/o Part-of-Speech graph & 77.3 $\pm$ 0.63 \\ 
SQLformer w/o dependency graph & 77.5 $\pm$ 0.72 \\ 
\hline
\textbf{SQLformer} (baseline) & \textbf{78.2 $\pm$ 0.75} \\ 
\hline
\end{tabular}}
\caption{EM accuracy (and $\pm$ 95\% confidence interval) of SQLformer ablation study on the Spider development set.}
\vspace{-.1in}
\label{tab:table5}
\end{table}

The results show that table and column selection has the biggest impact, with a performance drop from 78.2\% to 72.3\% when removed. This highlights the importance of schema-question linking. Removing the dependency graph and part-of-speech encodings leads to smaller decreases of 0.7\% and 0.9\%, respectively. Using an LSTM-based decoder from \cite{yin_syntactic_2017} instead of a Transformer-based one decreases performance by 4\%, demonstrating the effectiveness of our approach.

\section{Conclusion}

In this work, we introduced SQLformer, a novel model for text-to-SQL translation that leverages an autoregressive Transformer-based approach. SQLformer uses a specially designed encoder to link questions and schema and utilizes pre-trained models for effective language representation. Its unique decoder integrates node adjacency, type, and previous action information, conditioned on top-selected tables, columns, and schema-aware question encoding. Notably, SQLformer outperformed competitive text-to-SQL baselines across six datasets, demonstrating state-of-the-art performance.

\section*{Limitations}

One of the main limitations of our work is its focus on the English language, as it is the language used by most publicly available datasets. A potential way to alleviate this is by using multi-language PLMs for processing the questions.

%\section*{Ethics Statement}

%This work complies with the ACL Ethics Policy.

%\section*{Acknowledgements}

%We thank the anonymous reviewers.

% Entries for the entire Anthology, followed by custom entries
\bibliography{SQLformer}

\newpage

\appendix
%\section{Appendix}

\section{Details on structural information types} \label{appendix:structural_info_types}

All types of structural information used for question graph construction are shown in Table \ref{tab:dep_tags} and Table \ref{tab:pos_tags}. In particular, Table \ref{tab:dep_tags} highlights the syntactic dependency tags using during parsing of questions and Table \ref{tab:pos_tags} summarizes the semantic part-of-speech tags. Moreover, all relationships used for database schema graph construction are listed in Table \ref{tab:schema_relations}.

\begin{table}[h]
\centering
\begin{tabular}{|>{\ttfamily}l|p{5cm}|}
\hline
\textbf{Tag} & \textbf{Description} \\ \hline
ADJ  & Adjective: describes a noun or pronoun. \\ \hline
ADV  & Adverb: modifies a verb, adjective, or another adverb. \\ \hline
INTJ & Interjection: expresses a spontaneous feeling or reaction. \\ \hline
NOUN & Noun: names a specific object or set of objects. \\ \hline
PROPN & Proper Noun: names specific individuals, places, organizations. \\ \hline
VERB & Verb: describes an action, occurrence, or state of being. \\ \hline
ADP  & Adposition: relates to other words, specifying relationships. \\ \hline
AUX  & Auxiliary: helps form verb tenses, moods, or voices. \\ \hline
CCONJ & Coordinating Conjunction: connects words, phrases, or clauses of equal rank. \\ \hline
DET  & Determiner: modifies a noun, indicating reference. \\ \hline
NUM  & Numeral: represents a number. \\ \hline
\end{tabular}
\caption{Types of part-of-speech tags used during question graph construction}
\label{tab:pos_tags}
\end{table}

\begin{table}[h]
\centering
\begin{tabular}{|>{\ttfamily}l|p{5cm}|}
\hline
\textbf{Tag} & \textbf{Description} \\ \hline
ACL & Clausal modifier of noun. \\ \hline
ADVCL & Adverbial clause modifier. \\ \hline
ADVMOD & Adverbial modifier. \\ \hline
AMOD & Adjectival modifier. \\ \hline
APPOS & Appositional modifier. \\ \hline
AUX & Auxiliary. \\ \hline
CC & Coordinating conjunction. \\ \hline
CCOMP & Clausal complement. \\ \hline
COMP & Compound. \\ \hline
CONJ & Conjunct. \\ \hline
COP & Copula. \\ \hline
CSUBJ & Clausal subject. \\ \hline
DET & Determiner. \\ \hline
IOBJ & Indirect object. \\ \hline
NMOD & Nominal modifier. \\ \hline
NSUBJ & Nominal subject. \\ \hline
NUMMOD & Numeric modifier. \\ \hline
OBJ & Object. \\ \hline
\end{tabular}
\caption{Types of dependency parsing tags used during question graph construction}
\label{tab:dep_tags}
\end{table}

\begin{table}[h]
\centering
\resizebox{1\hsize}{!}{
\begin{tabular}{llll}
\toprule
Source node $x$ & Target node $y$ & Relationship & Description \\
\midrule
Table & Column & \textsc{Has-Column} & Column y belongs to the table x. \\
Column & Table & \textsc{Is-Primary-Key} & The column x is primary key of table y. \\
Column & Column & \textsc{Is-Foreign-Key} & Column x is the foreign key of column y. \\
Column & Literal & \textsc{Column-Type} & The column x has type y. \\
 \bottomrule
\end{tabular}
}
\caption{Summary of structural information types used in SQLformer during database schema graph construction.}
\label{tab:schema_relations}
\end{table}

\section{Analysis on training and inference efficiency} \label{appendix:train_inference_efficiency}

Table \ref{tab:table_tr_in_time} presents a comparative analysis of training and inference times between LSTM-based methods and SQLformer. Specifically, the average training time for every 50 iterations is calculated for both types of methods. The findings indicate that SQLformer achieves a training speed that is approximately four times faster and an inference speed that is 1.2 times faster than that of the LSTM-based methods.

\begin{table}[H]
\renewcommand\arraystretch{1.}
\centering
\setlength{\tabcolsep}{1mm}{
\small
\resizebox{1\columnwidth}{!}{
\begin{tabular}{=l|+c+c|+c+c|+c+c}
\hline
\multirow{2}{*}{\textbf{Method}}     & \multicolumn{2}{c}{\textbf{Spider}} & \multicolumn{2}{c}{\textbf{SParC}} & \multicolumn{2}{c}{\textbf{CoSQL}} \\ 
\cline{2-7}
& \textbf{Tr} & \textbf{In} & \textbf{Tr} & \textbf{In} & \textbf{Tr} & \textbf{In}  \\
\hline
LSTM & 203.1 & 19.3 & 174.2 & 18.5 & 162.8 & 19.7 \\ 
\hline
SQLformer & 52.9 & 16.2 & 67.4 & 15.6 & 53.7 & 15.8 \\ 
\hline
\end{tabular}}}
\caption{Training (Tr) and inference (In) efficiency comparison between LSTM-based approaches and SQLformer. Training efficiency is calculated as the average training time in seconds from 50 iterations. Inference efficiency is calculated as seconds per 100 queries.}
\vspace{-.1in}
\label{tab:table_tr_in_time}
\end{table}

\section{Details on the decoder architecture} \label{appendix:decoder_arch}

\begin{figure}[h]
\centering
\includegraphics[width=0.8\columnwidth]{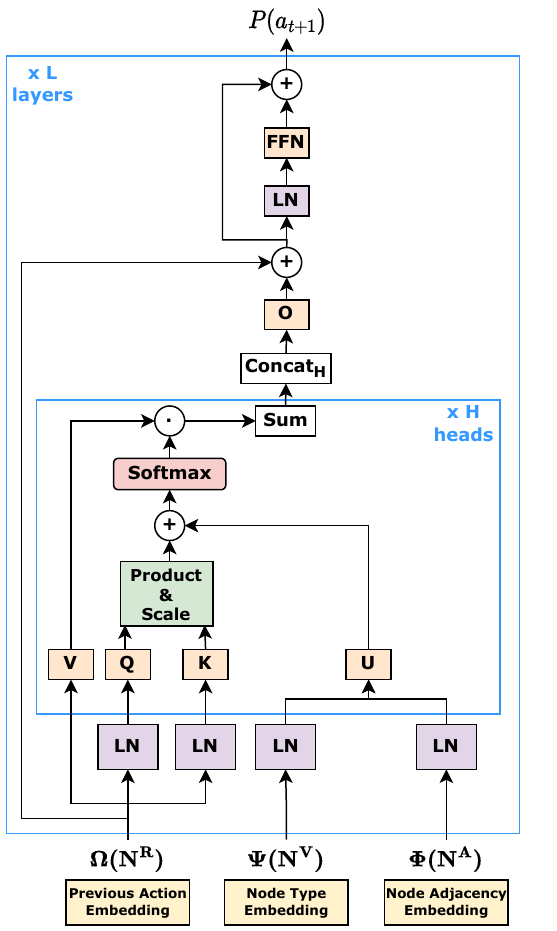}
\caption{Overview of the SQLformer decoder architecture.}
\vspace{-.1in}
\label{fig:app_decoder_arch}
\end{figure}

In the SQLformer decoder (Figure \ref{fig:app_decoder_arch}), the inputs are the node adjacencies and types in the current timestep of the generation process, as well as the previous action embedding. The node type and adjacency embeddings are integrated with the previous action embedding into the aggregation process of the MHA mechanism as a bias term. The node embedding is then transformed through a series of $L$ decoding layers with $H$ heads. The final representation is used to generate the probability distribution of actions to take in the next timestep.

\section{Summary of best hyperparameters used for SQLformer training} \label{appendix:hyperparameter_list}

For training SQLformer, we find the best set of hyperparameters on a randomly sampled subset of 10\% queries from the Spider dev split. Specifically, we find the best maximum previous AST nodes in the BFS ordering to be 30, and maximum training steps as 20,000. The number of layers for the encoder and decoder are both set to 6 and number of heads is 8. The dimensionality of the encoder and the decoder are set to 512. $k_{1}$ and $k_{2}$ are set to 10. The embedding sizes for tables and columns are set to 512. The node adjacency, node type and action embeddings sizes are 512. The output MLP for generating the next output action during decoding has 2 layers and hidden dimensionality of 512.

\section{Dataset details} \label{appendix:datasets}

For our experiments we use the (1) Spider dataset \cite{yu_spider_2019}, a large-scale cross-domain text-to-SQL benchmark. This dataset also incorporates multiple text-to-SQL datasets. The Spider dataset contains 8,659 training examples of question and SQL query pairs, 1,034 development (dev) examples and 2,147 test examples, spanning 300 complex databases across 138 different domains. Also, we run experiments on context-dependent settings with the (2) SParC and (3) CoSQL datasets, as well as zero-shot domain generalization performance on (4) Spider-DK, (5) Spider-SYN and (6) Spider-Realistic benchmarks. 

\section{Impact of Number of Selected Top Tables and Columns} \label{appendix:impact_k1_k2}

In this section, we analyze the impact of selecting different numbers of top tables and columns on the performance of SQLformer. The performance is measured using EM accuracy on the Spider development set. Table \ref{tab:table_impact_k1_k2} summarizes the results of varying the number of selected tables (\(k_1\)) and columns (\(k_2\)). As shown in the table, selecting more tables and columns generally improves the EM accuracy. However, this improvement comes with diminishing returns, indicating a trade-off between the number of selected schema elements and the model's complexity and efficiency.

\begin{table}[H]
\resizebox{1\columnwidth}{!}{
\begin{tabular}{c|c|c}
\hline
\textbf{\# tables (\(k_1\))} & \textbf{\# columns (\(k_2\))} & \textbf{EM accuracy (\%)} \\ 
\hline
5 & 5 & 73.1 \\ 
10 & 5 & 75.2 \\ 
5 & 10 & 76.7 \\ 
10 & 10 & 78.2 \\ 
\hline
\end{tabular}}
\caption{EM accuracy of SQLformer with varying numbers of top selected tables and columns.}
\vspace{-.1in}
\label{tab:table_impact_k1_k2}
\end{table}

These results demonstrate that while including more tables and columns can enhance performance, the gains are not linear and should be balanced against computational efficiency. Adjusting the number of top selected tables and columns can be a critical hyperparameter for optimizing performance in different application scenarios.

\end{document}